\documentclass{article}
\usepackage{ijcai22}

\usepackage{times}

\usepackage{soul}
\usepackage{url}
\usepackage[hidelinks]{hyperref}
\usepackage[utf8]{inputenc}
\usepackage[small]{caption}
\usepackage{graphicx}
\usepackage{amsmath}
\usepackage{booktabs}
\usepackage{comment}
\usepackage[caption=false]{subfig}
\usepackage{tikz}
\usepackage[english]{babel}
\usepackage{floatrow}
\title{Evolving-to-Learn Reinforcement Learning Tasks with Spiking Neural Networks}

\author{
J. Lu\footnote{Contact Author}\and
J.\,J. Hagenaars \and
G.\,C.\,H.\,E. de Croon
\affiliations
Micro Air Vehicle Laboratory, Delft University of Technology\\
}

\begin{document}

\maketitle

\begin{abstract}

Inspired by the natural nervous system, synaptic plasticity rules are applied to train spiking neural networks with local information, making them suitable for online learning on neuromorphic hardware. However, when such rules are implemented to learn different new tasks, they usually require a significant amount of work on task-dependent fine-tuning. This paper aims to make this process easier by employing an evolutionary algorithm that evolves suitable synaptic plasticity rules for the task at hand. More specifically, we provide a set of various local signals, a set of mathematical operators, and a global reward signal, after which a Cartesian genetic programming process finds an optimal learning rule from these components. Using this approach, we find learning rules that successfully solve an XOR and cart-pole task, and discover new learning rules that outperform the baseline rules from literature.

\end{abstract}

\section{Introduction}

In the 1990s, research progress in the field of neural dynamics in biological nervous systems promoted the development of spiking neural networks (SNNs)~\cite{Maas1997NetworksOS}. SNNs are a special type of artificial neural network (ANN) composed of biologically more realistic computational units. Similar to the communication between biological neurons, information transfer in SNNs is performed via spikes. A spike is a discrete event that occurs when the membrane potential of a neuron exceeds the threshold~\cite{Neuronal_Dynamics}. As a result, neurons in SNNs are sparsely activated, giving these networks the potential to perform computations more energy-efficiently compared to conventional ANNs, which communicate with dense, continuous values~\cite{Maas1997NetworksOS,Stone2016PrinciplesON}. For this reason, SNNs are promising for power-limited applications. Additionally, the recent development in neuromorphic chips has paved the way for the hardware implementation of SNNs~\cite{Davies2018LoihiAN,DeBole2019TrueNorthAF}, and they have been trained successfully with methods such as evolution~\cite{Howard2014EvolvingSN,Hagenaars2020EvolvedNC} or backpropagation of surrogate gradients~\cite{Shrestha2018SLAYERSL,Neftci2019SurrogateGL}. Nevertheless, these algorithms, making use of a lot of non-local information, are not particularly suitable for on-chip learning in neuromorphic hardware.

Efficient on-chip learning of SNNs can be realized through synaptic plasticity rules like Hebb's rule~\cite{hebb} or spike timing-dependent plasticity (STDP)~\cite{Markram}, or improved variants like Oja's rule~\cite{oja} and triplet STDP~\cite{Pfister2006TripletsOS}. Some of these learning rules were derived from experimental data~\cite{e2lbg1}, while others were derived to optimize some metric, like the efficiency of information transmission~\cite{Toyoizumi2005GeneralizedBR}. Most research takes both factors into account in the derivation process of the learning rule, but how much each aspect is relied on is rarely specified. Additionally, the discovery of alternative learning rules is usually a procedure of trial-and-error, which can be a time-consuming task. To solve these issues, Jordan et al.~\shortcite{e2l} proposed an evolutionary algorithm to discover the mathematical expressions of synaptic plasticity rules. Genetic programming was applied to evolve the rules to train SNNs on diverse tasks. The evolutionary search successfully (re)discovered existing solutions with high performance, and identified the essential terms for training. In some tasks, the algorithm also evolved new learning rules with competitive performance. However, since this work was the first attempt to evolve symbolic synaptic plasticity rules for SNNs, the tasks selected for the training were relatively simple and the search space was confined to the components present in existing solutions.

As an extension of \cite{e2l}, we will implement the same algorithm to evolve synaptic plasticity rules for training SNNs, targeting reinforcement learning tasks with increased task complexity and an expanded evolutionary search space. The objective is to evolve synaptic plasticity rules for solving reinforcement tasks, identify the critical terms for successful learning, and discover new learning rules with comparable or better performance. 

This paper will start with an overview of related work and the methodology in Section \ref{sec:related_works} and Section \ref{sec:methodology}, respectively. Next, Section \ref{sec:experiment} will present the results of the experiments. We will end with a discussion of the experiments and results in Section \ref{sec:discussion}.  

\section{Related Work}
\label{sec:related_works}

The idea of evolving synaptic plasticity rules was first applied to ANNs. To improve the learning capability of ANNs with biologically plausible models, Bengio et al.~\shortcite{Bengio1991LearningAS} optimized the parameters of synaptic plasticity rules using gradient descent and a genetic algorithm. Later, the generality of this approach was demonstrated by applying it to learn simple but diverse tasks: a biological circuit, a Boolean function, and various classification tasks~\cite{BengioS}. Following a similar approach, Niv et al.~\shortcite{nivy} and Stanley et al.~\shortcite{stanley} evolved learning rules to train ANNs to perform reinforcement learning tasks and automatic control tasks with adaptive environments, respectively. In all these works, the optimization of the learning rules is based on a parametric function, which is usually an existing synaptic plasticity rule such as Hebb's rule. The optimizer only learns the coefficients of the function. The main advantage of this approach is that fixing the format of the learning rule constrains the search space and reduces the computational effort. However, the robustness and generality of this approach are compromised since variables outside the parametric function are all excluded.

A more general method is to replace the symbolic learning rule with an ANN and optimize it for better learning performance. The adaptive HyperNeat algorithm is a meta-learning method that evolves the connection pattern and weights of the ANN which encodes the learning rule~\cite{hyperneat}. This approach uses neuronal states such as neuron traces and spike activities as inputs to the ANN, which outputs the weight update of the synapse in question. Recently, a similar approach has been implemented in the neuromorphic hardware to optimize the training of SNNs in basic reinforcement learning tasks~\cite{bohnstingl}. This approach simplifies the adaptive HyperNeat algorithm and only evolves the connection weights of the ANN. Both works compared their approaches with methods that optimize a parametric learning rule on the same tasks, and it was found that optimizing the learning rule encoded by an ANN is more robust to tasks with changing environments. The drawback of this approach is also obvious. Since the learning rules are expressed by ANNs, which are hardly interpretable, it is difficult to analyze the learning behaviour and generalize the evolved learning rules to learn other tasks. 

With the development of computational neuroscience and SNNs, various synaptic plasticity rules have been derived for learning diverse tasks~\cite{Markram,florian_reinforcement_2007}. To come up with more general rules that can be applied to a broader array of problems, researchers have proposed meta-learning approaches that evolve and evaluate the symbolic expressions of the synaptic plasticity rules. Different from the previous methods that have a fixed function, the coefficients, operators, and operands of the learning rule are all genomes to be evolved in this approach. Jordan et al.~\shortcite{e2l} implemented Cartesian genetic programming (CGP) to evolve synaptic plasticity rules for learning reward-driven, error-driven, and correlation tasks. Confavreux et al.~\shortcite{Confavreux} successfully applied covariance matrix adaptation evolution strategy to rediscover Oja’s rule and an anti-Hebbian rule. Compared to the optimization algorithms discussed in the previous paragraphs, these approaches evolve the entire expression of the learning rule. They have a larger search space than methods that only evolve coefficients. Therefore, they are also more robust and flexible. Furthermore, the result of the evolution is a mathematical expression of the learning rule, which makes interpretation of learning behaviour easier. For these reasons, we will follow a similar approach in this paper. 

\section{Methodology}
\label{sec:methodology}

\subsection{Neuron Model}

There are a number of neuron models with varying levels of complexity and fidelity \cite{hh_model,LF_model}. Because the evolution of learning rules for training SNNs is computationally very expensive, a neuron model with low complexity is favorable. Furthermore, high fidelity regarding the biological neural system is not an essential factor for succesful evolution. For these reasons, the leaky integrate-and-fire (LIF) model is sufficient for the spiking neuron simulation here. Analogous to an RC circuit, the neuron membrane can be considered as a capacitor and the membrane potential $U$ varies with the current $I$ flowing into it, which is the summation process. When the membrane potential reaches the spiking threshold, the neuron releases a spike $S$ and the membrane potential drops to the equilibrium or resting value. The leaky behavior describes the decay of the potential due to ion diffusion. The LIF model can be expressed mathematically as follows:
\begin{align}
    U_{i}^{k} &= (1 - S_{i}^{k-1})\alpha U_{i}^{k-1} + (1-\alpha)I_{i}^{k}\label{eq:lif1}\\
    I_{i}^{k} &= \sum_{j}w_{ij}S_{j}^{k}
    \label{eq:lif2}
\end{align}

\noindent where $i$, $j$ indicate the post- and presynaptic neurons respectively, and $k$ indicates the simulation timestep. The constant $\alpha$ is the membrane decay and $w$ is the synaptic weight.

\subsection{Reward Modulated Spike Timing Dependent Plasticity (R-STDP)}

STDP is a learning rule fitted to experimental data of biological neurons~\cite{Markram}, and variants of STDP have been widely applied for training SNNs. The biological foundation is Hebb's theory, which is summarized as ``cells that fire together wire together"~\cite{hebb}. It was later extended with the long-term potentiation and long-term depression phenomena discovered in the rabbit hippocampus~\cite{lomo}. The long term potentiation (LTP) phenomenon shows that a long-term increase in synaptic strength occurs when the postsynaptic neuron fires after the presynaptic spike in a certain time window. Oppositely, a long-term decrease takes place when the postsynaptic neuron fires before the presynaptic spike in a certain time window. Thus, in the basic STDP learning rule, synapse strength change is a function of the time difference between the pre- and postsynaptic spikes. In some variations of the basic STDP, the eligibility trace of the pre- and postsynaptic neurons or synapses are used to represent the time difference of spikes. To further imitate the reward driven learning behaviour in nature, reward-modulated STDP (R-STDP) introduces a reward signal in the learning rule, making it suitable for training SNNs to perform reinforcement learning tasks. An example is the MSTDPET rule proposed by Florian~\shortcite{florian_reinforcement_2007}:
\begin{align}
    w_{ij}(t+\Delta t) &= w_{ij}(t)+\gamma R(t+\Delta t)E_{ij}(t+\Delta t) \label{eq:MSTDPET} \\
    E_{ij}(t+\Delta t) &= E_{ij}(t)e^{-\Delta t/\tau_{z}} + \xi_{ij}(t) \label{eq:MSTDPETz} \\
    \xi_{ij}(t) &= X_j^+(t)S_i(t) + X_i^-(t)S_j(t) \\
    X_j^+(t) &= X_j^+(t - \Delta t)e^{-\Delta t/\tau_{+}} + A_+S_j(t) \\
    X_i^-(t) &= X_i^-(t - \Delta t)e^{-\Delta t/\tau_{-}} + A_-S_i(t) \label{eq:MSTDPETlast}
\end{align}

\noindent where $\gamma$ is the learning rate and $\Delta t$ is the simulation timestep. The reward $R$ is updated at each timestep. The eligibility trace at each synapse $E_{ij}$ is a low-pass filter over synaptic activity $\xi_{ij}$, which has a pre-before-post ($+$) and a post-before-pre ($-$) term. All $\tau$s represent decay time constants. Constants $A_\pm$ scale the contributions of spikes $S$ to the neuron activity traces $X^\pm$.

\subsection{Evolving-to-learn}

\begin{figure}
    \centering
    \includegraphics[trim={0cm 5cm 0cm 5cm},width=1.0\columnwidth]{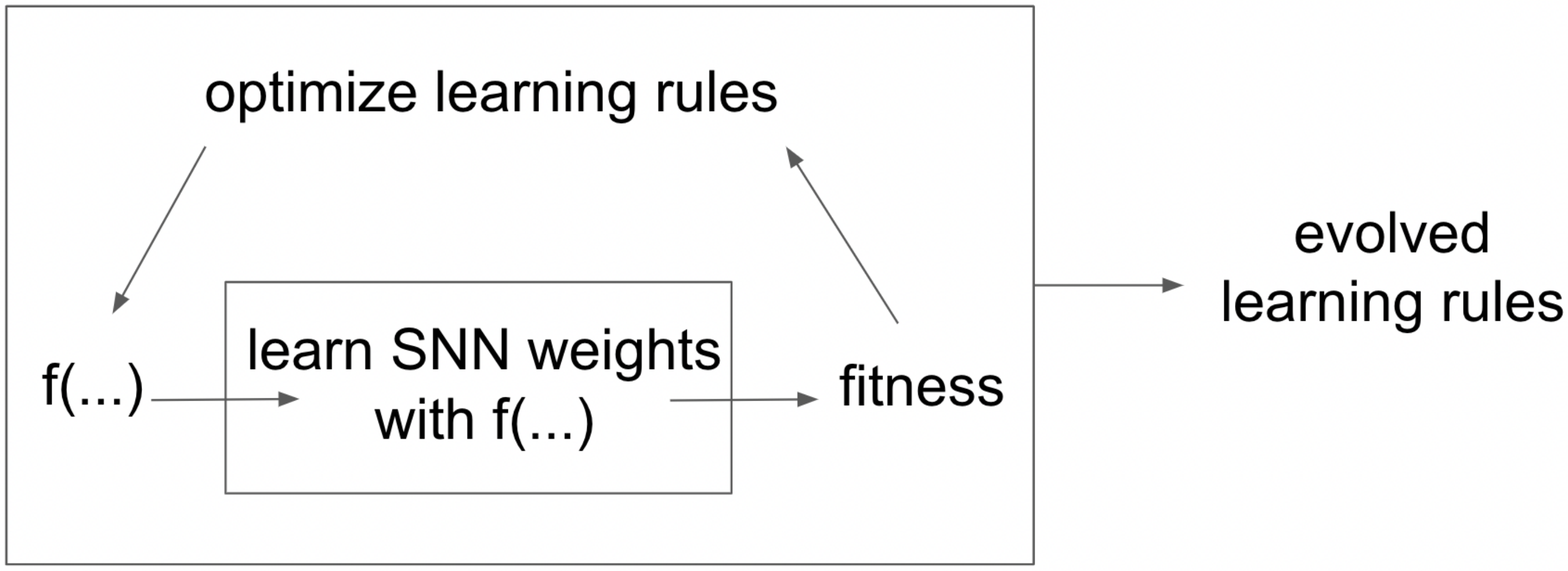}
    \caption{\textbf{Two learning loops of the evolving-to-learn algorithm.} The outer loop evolves the learning rule $f(...)$ to learn the task in the inner loop. The output of the inner loop is the fitness of the learning rule, which is used to select better learning rules.}
    \label{fig:e2l_loops}
\end{figure}

The evolving-to-learn algorithm proposed by Jordan et al.~\shortcite{e2l} can be considered as an optimization-based meta-learning algorithm with two loops, as visualized in \autoref{fig:e2l_loops}. The outer loop is an evolutionary search for the mathematical expressions of learning rules, which will be performed with CGP~\cite{cgp}. The learning rules are represented by an indexed 2D Cartesian graph. The graph has a fixed number of input, output, and internal nodes. Each internal node has a string of three integers: the first integer indicates the operator and the other two integers indicate the index of nodes to be connected. The output of the internal nodes can also be used as input to other internal nodes. 

\begin{figure}
    \centering
    \includegraphics[trim={1cm 0cm 0cm 0cm},width=\columnwidth]{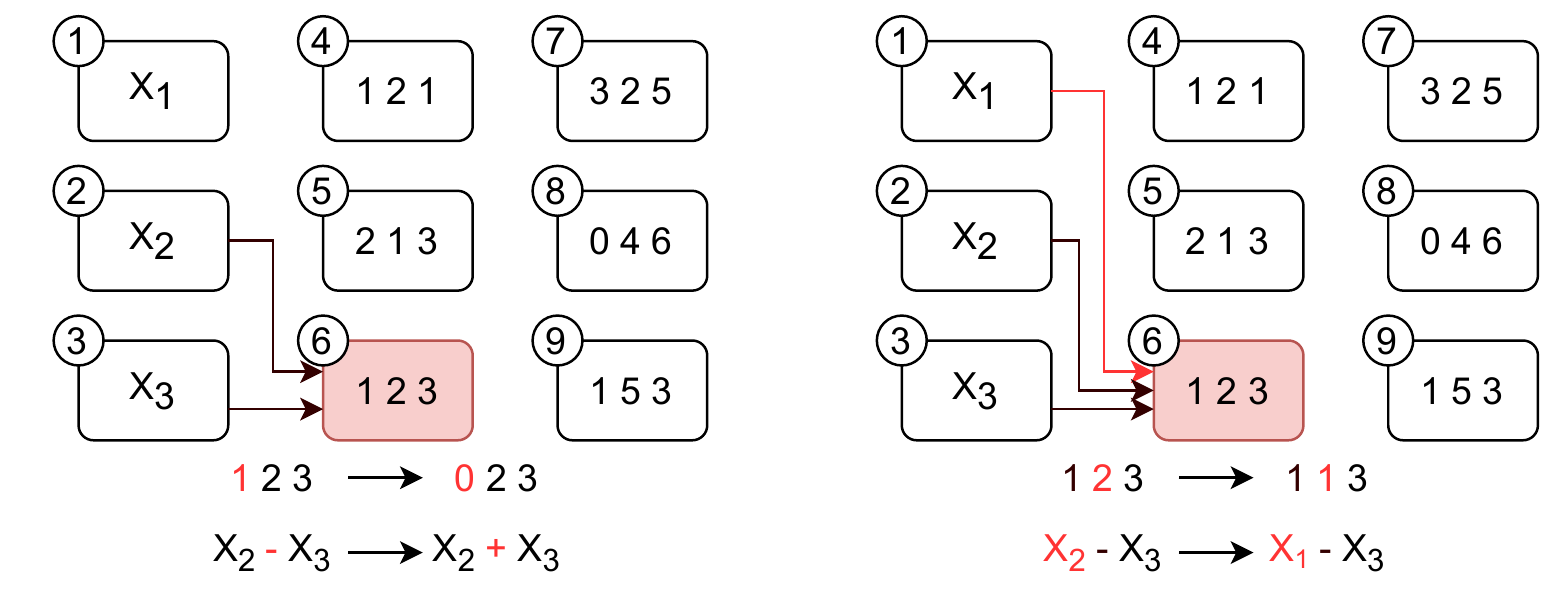}
    \caption{\textbf{An example of the mutation in CGP.} The left graph shows the mutation in the operator. The first index 1 is replaced by 0, the minus operator is mutated to a plus sign. The right graph shows the mutation in the operand. The index corresponding to the input $x_{2}$ is mutated to the index of the input $x_{1}$.}
    \label{fig:CGP}
\end{figure}

At the beginning of the evolution, the indexed graphs are randomly initialized. After each generation, the $\mu$ best solutions will be selected as parents and an offspring of size $\lambda$ will be generated from the parents through mutation. Mutation can take place in both the operators and the connection of the nodes. An example of CGP with three columns and three rows is shown in Figure~\ref{fig:CGP}. 

Learning rules generated by CGP will then be used to train SNNs to perform a particular task in the inner loop. The performance of the inner loop is in turn the fitness of the outer loop for the evolution. Summarized, the algorithm can be described by the following steps:\footnote{Code will be made open-source upon publication.}

\begin{enumerate}
    \item Define the task to be learned for the inner loop and the fitness function.
    \item Design the SNN architecture and determine encoding-decoding methods. 
    \item Specify the input signals for the evolutionary search.
    \item Perform the evolution to discover learning rules with high fitness.
\end{enumerate}

The same procedure will be implemented in our experiments. However, different from the work of Jordan et al.~\shortcite{e2l} which used only the input signals of baseline learning rules, we also included neuronal states such as spikes and neuron traces, in order to give the evolution more freedom. Hyperparameters such as the membrane decay and threshold of spike are not included in the evolution; they are pre-tuned with the baseline learning rule and remain fixed.

\section{Experiments}
\label{sec:experiment}

\subsection{XOR Classification}

We first test the proposed method on the elementary XOR classification task. Each XOR input is encoded as a spike train of $500$ timesteps. $50$ spikes are randomly distributed within the interval, generating two distinct patterns for $0$ and $1$ respectively. At the beginning of each learning epoch, different sets of input patterns are generated as training and test samples. The SNN has three layers of LIF neurons: $2$ input neurons, $20$ hidden neurons, and a single output neuron. The classification depends on the total number of output spikes. The output will be $1$ if the output neuron spikes more than the average output spikes of one learning cycle; else, the output will be $0$.

As a baseline learning rule, we use the MSTDPET rule~\cite{florian_reinforcement_2007}, as given in Equations~\ref{eq:MSTDPET}-\ref{eq:MSTDPETlast}. Different from the Urbanczik and Senn's rule used in~\cite{e2l}, MSTDPET updates connection weights every timestep of the simulation. Therefore, the reward should also be assigned to the network at each timestep. If the correct output is $1$, a reward of $1$ will be assigned to the synapse for each output spike released. If the correct output is $0$, the reward is $-1$ for each output spike released. In the absence of output spikes, the reward for both cases is $0$. 

As discussed before, the signals available to the baseline learning rule will be included in the evolution, which in this case are $E_{ij}$ and $R$. To expand the search space and endow more freedom to the evolutionary search, neuronal states that play a role in biological synaptic plasticity are also made available to the evolution: pre- and postsynaptic spikes $S_{i}$ and $S_{j}$, and pre- and postsynaptic neuron traces with two different decays $e_{i1}$, $e_{i2}$, $e_{j1}$ and $e_{j2}$. The general form of the learning rule is then expressed as follows:
\begin{equation}
    \Delta w_{ij} = \gamma f(E_{ij}, R, S_{i}, S_{j}, e_{i1}, e_{i2}, e_{j1}, e_{j2})
    \label{eq:general1}
\end{equation}

Since the weight initialization is performed randomly, there can be cases that the initialization is already near an optimal solution, resulting in good performance, and also cases where the initialization is unfavorable for learning. Thus, each individual performs three trials of the experiment in the evolutionary search. The fitness function is defined to be the average of the end test accuracy of the three trials. 

\begin{figure}
    \centering
    \includegraphics[trim={1cm 0cm 1cm 0cm},width=1.0\columnwidth]{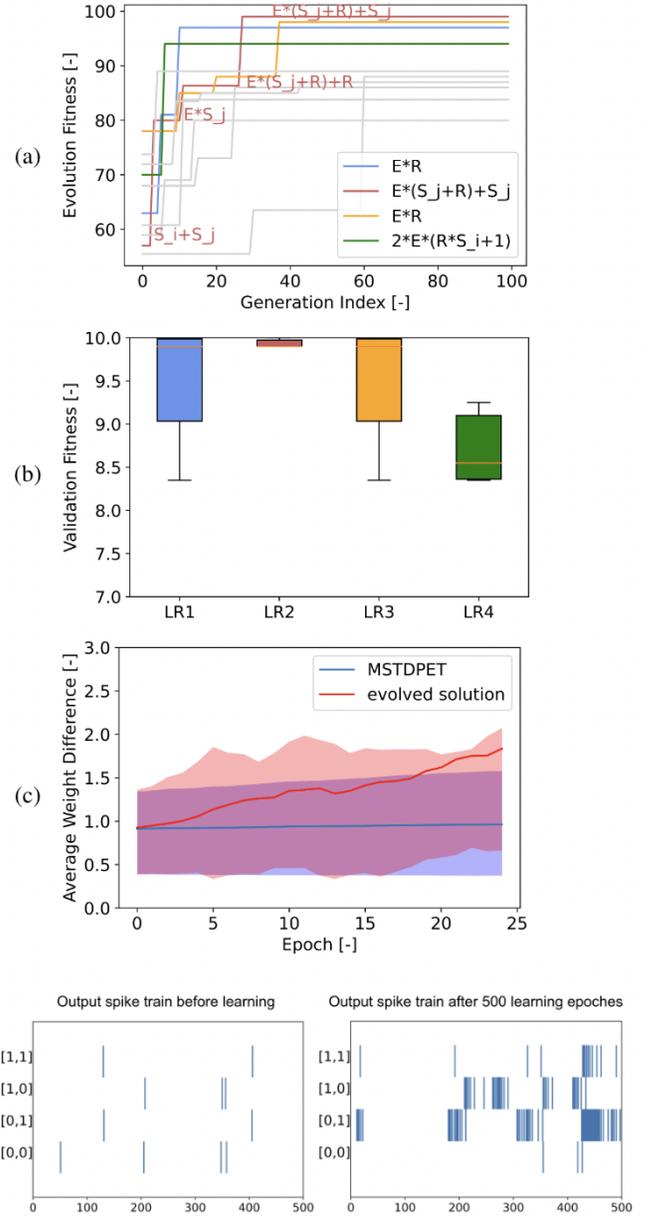}
    \caption{\textbf{Results of the evolution for learning the XOR task.} (a) The learning curve of the evolutionary search shows the fitness of the best solution at each generation. The grey curves are trials that failed to discover solutions with performance comparable to the baseline learning rule; the others are successful trials. The evolutionary process is also shown in the plot for the best learning trial. (b) The final evolved learning rules of the successful trials were validated with $50$ random learning cycles. The average test accuracy at the end of the learning with the inter-quartile range (IQR) is plotted. (c) The plot shows the average variation of the difference between the weights connected to the two input neurons for both the baseline learning rule and the best evolved solution. The shaded area represents the IQR. (d) The output spike trains before and after learning with the evolved learning rule are visualized.}
\label{fig:result1}
\end{figure}

The results of $10$ evolutionary runs are shown in Figure~\ref{fig:result1}(a-b). It shows that two of the learning cycles successfully evolved the MSTDPET learning rule within $100$ generations. Besides, it can be noticed that the term $ER$ appeared in all of the evolved learning rules with high fitness, indicating that it is critical for learning the XOR task. This can be explained as follows. With temporal encoding, the two input neurons will spike simultaneously for pattern $0$. Synapses between a postsynaptic neuron and the two input neurons will have an identical $E$ at each timestep. Receiving a negative reward for each output spike released, the synapse weights will also be decreased by the same amount. However, when the label of the input is $1$, the two input spike trains will be different. If the synapse connection is strong, there is a larger possibility that the postsynaptic neuron will spike right after the presynaptic spike resulting in a larger synapse trace. Oppositely, connections with smaller weights tend to have smaller synaptic traces. Thus, after a number of learning cycles, strong connections will be stronger and the weak get weaker, which is a pattern favorable to solving the XOR task. 

Aside from rediscovering the baseline learning rule, a learning rule that achieved a better performance was discovered, which has the expression $E(S_{j} + R) + S_{j}$. In addition to the synapse trace and reward, the presynaptic spike is also included in the learning rule. Similar to the MSTDPET rule, the evolved rule learns the XOR task by increasing the strength difference between the synapses connected to the two input neurons. With the baseline learning rule, weights will only be updated when the postsynaptic neuron spikes. The evolved rule also updates weights when the presynaptic neuron spikes. For pattern $0$, weights of synapses connected to both input neurons will increase identically when there is only a presynaptic spike and decrease when there is only an output spike. For pattern $1$, both the presynaptic spike and output spike will increase the strength difference between the synapses connected to two input neurons, thereby speeding up the learning as shown in Figure~\ref{fig:result1}(c). Figure~\ref{fig:result1}(d) demonstrates that after $500$ learning epochs with the evolved learning rule, the output spikes rate for input $[0,1]$ and $[1,0]$ is significantly larger.

\subsection{Cart-pole Task}

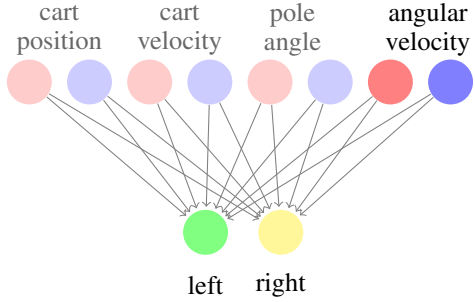
\begin{figure}
\centering
\begin{tikzpicture}[shorten >=1pt,->,draw=black!50, node distance=\layersep]
    \tikzstyle{every pin edge}=[<-,shorten <=1pt]
    \tikzstyle{neuron}=[circle,fill=black!25,minimum size=17pt,inner sep=0pt]
    \tikzstyle{left neuron}=[neuron, fill=green!50];
    \tikzstyle{positive neuron}=[neuron, fill=red!50, fill opacity=0.4];
    \tikzstyle{negative neuron}=[neuron, fill=blue!50, fill opacity=0.4];
    \tikzstyle{positive neuron1}=[neuron, fill=red!50];
    \tikzstyle{negative neuron1}=[neuron, fill=blue!50];
    \tikzstyle{right neuron}=[neuron, fill=yellow!50];
    \tikzstyle{annot} = [text width=3cm, text centered]
    \foreach \name / \x in {2,4,6}
            \node[negative neuron] (I-\name) at (0.8*\x,0) {};
        
    \foreach \name / \x in {1,3,5}
            \node[positive neuron] (I-\name) at (0.8*\x,0) {};

    \foreach \name / \x in {8}
            \node[negative neuron1] (I-\name) at (0.8*\x,0) {};

    \foreach \name / \x in {7}
            \node[positive neuron1] (I-\name) at (0.8*\x,0) {};            

    \foreach \name / \x in {1}
        \path[xshift=0.25\columnwidth]
            node[left neuron] (H-\name) at (\x cm,-2) {};

    \foreach \name / \x in {2}
        \path[xshift=0.25\columnwidth]
            node[right neuron] (H-\name) at (\x cm,-2) {};

    \foreach \source in {1,...,8}
        \foreach \dest in {1,2}
            \path (I-\source) edge (H-\dest);
    \node[annot,opacity=0.6,above of=I-1, node distance=0.7cm, xshift=0.4cm] (hl) {cart\\position};
    \node[annot,opacity=0.6,above of=I-3, node distance=0.7cm, xshift=0.4cm] (hl) {cart\\velocity};
    \node[annot,opacity=0.6,above of=I-5, node distance=0.7cm, xshift=0.3cm] (hl) {pole\\angle};
    \node[annot,above of=I-7, node distance=0.7cm, xshift=0.5cm] (hl) {angular\\velocity};    
    \node[annot,below of=H-1, node distance=0.7cm] (hl) {left};  
    \node[annot,below of=H-2, node distance=0.7cm] (hl) {right};
\end{tikzpicture}
\caption{\textbf{The architecture of the SNN for the cart-pole task.} Each blue circle represents five neurons for negative values and each red circle represents five neurons for positive values. Only the input neurons corresponding to the angular velocity are used in the final experiment.}
\label{fig:snn_archi}
\end{figure}

Next, the same evolving-to-learn approach was implemented to learn plasticity rules for training SNNs to perform the cart-pole task. The CartPole environment of OpenAI Gym\footnote{\url{https://gym.openai.com/envs/CartPole-v1/}} was used, where four observation states were available: cart position, cart velocity, pole angle, and pole angular velocity. The actions of the system are discrete: a force exerted on the cart towards the left or right. The SNN has two layers: an input layer of pairs of Poisson encoders and $2$ output neurons. Each pair of Poisson encoders has $5$ neurons for positive values and $5$ neurons for negative values. The Poisson encoders convert the absolute value of the input state into a spike train with $50$ network simulation timesteps. One of the binary encoders will be activated according to the sign of the state. An episode is a simulation cycle of the cart-pole with a number of environment timesteps. For each environment timestep, the model will receive a set of observation states. After being encoded into spike trains, the SNN will go through the network simulation. At the end of the network simulation, the system will take an action based on the output of the model. The $2$ output neurons control the action of the cart. If the output neuron corresponding to the left action spikes more than the other neuron, a force towards the left will be applied to the cart. Otherwise, a force towards the right will be exerted. The architecture of the SNN is visualized in Figure~\ref{fig:snn_archi}.

After the system takes an action, the model will receive a reward from the environment. If the action is to the left and causes the pole to move to the center, the synapses connected to the left output neuron will receive a positive reward while the other synapses will receive an opposite reward. If the action to the left causes the pole to fall, the connection with the output neuron corresponding to the left will receive a negative reward, and the other connections will receive a positive reward. 

The baseline learning rule for learning the cart-pole task is the multiplicative R-STDP rule used in~\cite{Shim2017BiologicallyIR} to learn robot control tasks:
\begin{equation}
    w_{ij}(t+\Delta t) = w_{ij}(t)+\gamma R(t+\Delta t)E_{ij}(t+\Delta t)w_{ij}(0)
    \label{eq:base_cartpole}   
\end{equation}
Different from MSTDPET, this learning rule involves the initial weight at the beginning of the learning window $w_{ij}(0)$ to boost the performance. However, the problem with applying this learning rule is that the reward has to be available at each network simulation timestep while the SNN model will only receive the feedback after taking an action at the end of the network simulation. To solve this issue, the neuronal states at each network timestep were stored during the simulation. After the action is taken, the weight will be updated with the reward and stored states. Next, it was noticed that the baseline learning rule failed to perform the cart-pole task stably. This is because in cases where the pole falls and pushes the cart to move fast, the output gets dominated by a large-magnitude state, leading to very-high-rate input spike trains. For this reason we modified the experiment: 1) instead of receiving all four states (cart position, cart velocity, pole angle, and pole angular velocity), the network receives only the pole angular velocity (most essential information); 2) the network weights were only updated during a certain amount of timesteps at the beginning of each episode.

Similar to the XOR task, pre- and postsynaptic neuron activities and traces are also included in the evolution along with the input variables of the baseline learning rule. The synaptic plasticity rule can be expressed as follows: 
\begin{equation}
    \Delta w_{ij} = \gamma f(E_{ij}, R, f_{i}, f_{j}, e_{i1}, e_{i2}, e_{j1}, e_{j2}, w_{ij}(0))
    \label{eq:general2}
\end{equation}
During the evolution, each candidate learning rule has three random trials of the task and each trial has $50$ episodes. The maximum lifespan is $100$ environment timesteps, and the episode will be terminated when the pole angle exceeds $15^\circ$. The fitness of the candidate learning rule is measured by the average time balance time of the last $5$ episodes.  

\begin{figure}[hbtp!]
    \centering
    \includegraphics[width=0.9\columnwidth]{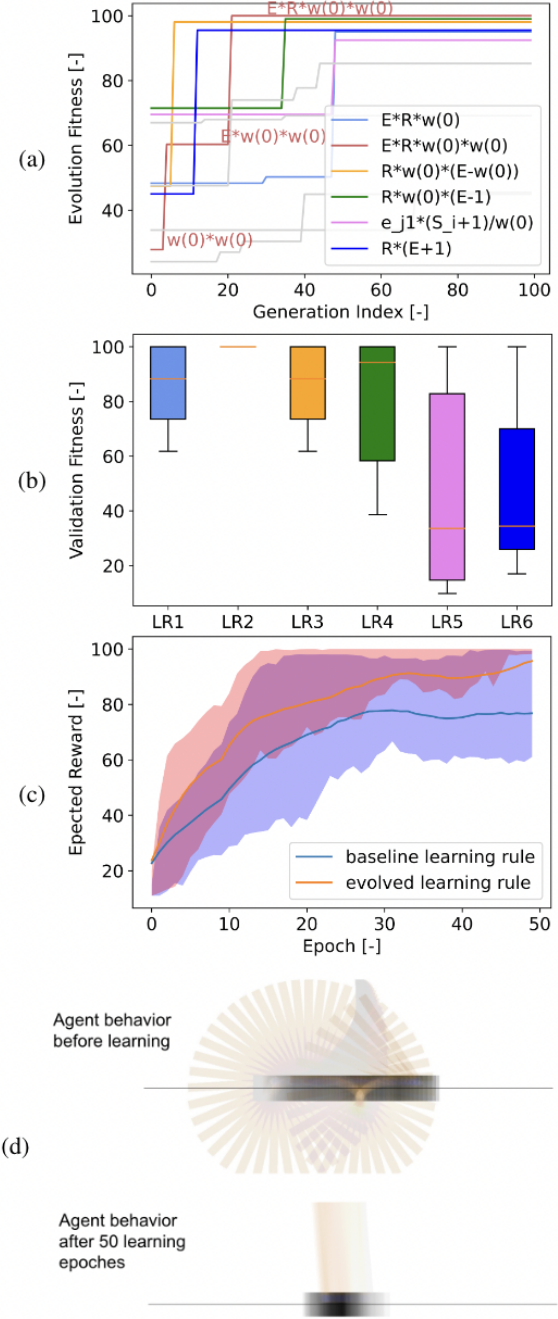}
\caption{\textbf{Results of the evolution for learning the cart-pole task.} (a) The learning curve of the evolutionary search shows the fitness of the best solution at each generation. The grey curves are trials that failed to discover solutions with comparable performance as the baseline learning rule, the others are successful trials. The evolution process is also shown in the plot for the best learning trial. (b) The final evolved learning rules of the successful trials were validated with $50$ random learning cycles. The average test accuracy at the end of the learning with IQR is plotted. (c) The plot shows the average learning curves of $50$ validation runs for both the baseline learning rule and the evolved learning rule. The shaded area represents the IQR. (d) The agent behaviour trajectory of one episode before and after learning with the evolved learning rule are visualized. To show the entire trajectory, the episode was not terminated when the pole angle exceeded $15^{\circ}$.}
\label{fig:result2}
\end{figure}

The results of $10$ evolution runs are shown in Figure \ref{fig:result2}(a-b). It can be noticed that one of the evolution runs resulted in the baseline learning rule exactly and another run evolved a similar expression with comparable fitness: $Rw(0)(E-1)$. For strong connections with large synaptic traces, the additional constant term is negligible and, therefore, has little influence on the learning performance. There are also two evolution runs that obtained high fitness during the evolution. However, they had an unstable learning performance and achieved a lower fitness in the validation. More importantly, the evolution also discovered a learning rule which obtained higher fitness and a more stable learning performance than other rules: $\Delta w_{ij}(t) = \gamma tR(t)E_{ij}(t)w_{ij}(0)^{2}$. Compared to the baseline learning rule, it includes all the inputs from the multiplicative R-STDP (Equation~\ref{eq:base_cartpole}), but the initial weight has a power of two. It boosts the learning performance compared to the baseline learning rule as shown in Figure~\ref{fig:result2}(c). When the strong connection results in the correct action and obtains a positive reward, the learning rule will further strengthen the connection pattern. However, when the strong connection results in a wrong action and obtains a negative reward, the learning rule will decrease the strong connection weights and increase the weak connection weights. The $w_{ij}(0)^{2}$ term will lead to the desired pattern faster. Figure~\ref{fig:result2}(d) demonstrates that the pole remained balanced for the entire episode after $50$ learning epochs with the evolved learning rule.

\section{Discussion}
\label{sec:discussion}

In this article, we applied an evolving-to-learn algorithm to evolve mathematical expressions of synaptic plasticity learning rules using local input signals. The algorithm successfully learned learning rules for training SNNs on the XOR and cart-pole task. As demonstrated by the experiment results, both the XOR and cart-pole experiments (re)discovered the baseline learning rules and identified the essential terms for solving the tasks. This approach allows for better generalization of the synaptic plasticity rules to other similar tasks, since the learning rule can be customized based on these essential terms. Meanwhile, compared to previous work~\cite{e2l,Confavreux}, we expanded the search space and solved a more complex problem in the form of the cart-pole task, further proving the generality of the algorithm and the flexibility of the evolutionary approach. Additionally, for both experiments we discovered a new learning rule with higher performance and stability than the baseline solution.  

During the experiments, the significance of the prior knowledge in the existing synaptic plasticity rules was revealed. Not only because the baseline learning rule provided a part of the input signals, but also because preliminary learning was necessary for tuning the hyperparameters of the network model to ensure that the model was active for solving the task. However, since the hyperparameters were fixed during the evolution, it may discourage the discovery of new learning rules which work with other sets of hyperparameters.

Moreover, there is a part of the experiment that is usually neglected by other works and other research on synaptic plasticity learning for SNNs. In this work, it was found that the design of the experiment has a large influence on the evolution of the learning rule. For instance, the encoding and decoding, the definition of the reward are all elements that are important to training the SNN model. If one of these elements is not properly defined, it can lead to low performance for both the evolution as well as the learning rules themselves. These design decisions are usually made through trial-and-error and iteration during the experiments. Therefore, although the evolving-to-learn algorithm reduces the workload for manually deriving and comparing different synaptic plasticity rules, there is still a large amount of work for designing the experiment properly. 

In conclusion, the evolving-to-learn algorithm is an effective tool for helping to understand the learning procedure of an existing synaptic plasticity rule and assisting the manual generalization of the learning rule. It is also capable of discovering and optimizing learning rules with a predefined experimental setup. However, due to the dependence on prior knowledge and the sensitivity to the design decisions, the algorithm is not yet able to fully automatically discover new learning rules for a random new task. One next step towards a higher level meta-learning of synaptic plasticity rules can be including an automatic tuning of the hyperparameters in the evolution, which may allow the discovery of more new learning rules with the corresponding hyperparameters.

\newpage
\bibliographystyle{named}
\bibliography{ijcai22}

\newpage
\appendix
\section{Hyperparameters}
\autoref{tab:hyperparameters}  lists the hyperparameters used in the SNNs and evolution for XOR and cart-pole tasks respectively.

\begin{table}[htp]
\caption{Hyperparameters for the XOR and cart-pole experiments.\label{tab:hyperparameters}}
\begin{tabular}{l|l|l}
\hline
\textbf{Hyperparameters}                                        & \textbf{XOR}                                            & \textbf{Cart-pole}                                      \\ \hline
number of spikes per  pattern ($N$)                               & 50                                                      & -                                                       \\
number of patterns per training epoch ($M$)                             & 4                                                       & -                                                       \\
number of episode per epoch ($N_{ep}$)                            & -                                                       & 100                                                     \\
simulation duration ($T$)                                         & 500                                                     & 50                                                      \\
environment duration ($T_{env}$)                                   & -                                                       & 100                                                     \\
number of training epoches ($N_e$)                               & 500                                                     & 50                                                      \\
SNN architecture                                                & {[}2, 20, 1{]}                                          & {[}10, 2{]}                                             \\
membrane potential decay ($\tau_{V}$)           & 10                                                      & 10                                                      \\
synapse trace decay ($\tau_{e}$)                & 100                                                     & 100                                                     \\
spike threshold ($\theta$)                         & 100                                                     & 1                                                       \\
activity decay ($\tau_+$, $\tau_-$) & 10                                                      & 2                                                       \\
pre- and postspike constants ($A_{+}$, $A_{-}$)                       & +1, -1                                                  & +1.5, -0.5                                              \\
reward ($R$)                                                      & +1, -1                                                  & +1, -0.5                                                \\
leanring rate ($\eta$)                             & 5 e-3                                 & e-5                                   \\
number of parents ($\mu$)                          & 20                                                      & 50                                                      \\
number of offsprings ($\lambda$)                   & 20                                                      & 50                                                      \\
number of generations ($N_{g}$)                                    & 100                                                     & 100                                                     \\
CGP architecture                                                & 2 $\times$ 12                              & 2 $\times$ 12                              \\
operators                                                       & + - $\times$ \textbackslash constantfloats & + - $\times$ \textbackslash constantfloats \\
mutation rate                                                   & 0.3                                                     & 0.3                                                     \\ \hline
\end{tabular}
\end{table}

\end{document}